\documentclass[runningheads]{llncs}
\usepackage[T1]{fontenc}
\usepackage{graphicx}
\usepackage{subfigure}
\usepackage{bbding}
\usepackage{framed,multirow}
\usepackage{amssymb}
\usepackage{latexsym}
\usepackage{amsmath}
\usepackage{url}
\usepackage{xcolor}
\usepackage{svg}
\usepackage{hyperref}
\usepackage{mathtools}
\usepackage{hyperref}
\hypersetup{urlcolor=blue}
\usepackage{bbding}
\usepackage[marginal]{footmisc}
\renewcommand{\thefootnote}{}

\begin{document}
\title{TSegFormer: 3D Tooth Segmentation in Intraoral Scans with Geometry Guided Transformer
%\thanks{Supported by organization x.}
}

\titlerunning{TSegFormer}

\author{Huimin Xiong\inst{1,2} \textsuperscript{*} \and
Kunle Li\inst{1} \textsuperscript{*} \and
Kaiyuan Tan\inst{1} \textsuperscript{*} \and 
Yang Feng\inst{4} \and
Joey Tianyi Zhou\inst{5,6} \and
Jin Hao \inst{7} \and
Haochao Ying \inst{3} \and
Jian Wu \inst{3} \and
Zuozhu Liu \inst{1,2(}\Envelope\inst{)}
  }
%index{Xiong, Huimin; Li, Kunle; Tan, Kaiyuan; Feng, Yang; Zhou, Joey Tianyi; Hao, Jin; Ying, Haochao; Wu, Jian; Liu, Zuozhu}

% 
\authorrunning{H. Xiong et al.}

\institute{
ZJU-UIUC Institute, Zhejiang University, Haining, 314400, China 
\email{zuozhuliu@intl.zju.edu.cn} \\ \and
Stomatology Hospital, School of Stomatology, 
Zhejiang University School of Medicine, Hangzhou, 310058, 
China \and
School of Public Health, Zhejiang University, Hangzhou, 310058, China \and
Angelalign Research Institute, Angel Align Inc., Shanghai, 200011, 
China \and
% A*STAR Centre for Frontier AI Research (CFAR), 138632, 
% Singapore \and
Centre for Frontier AI Research (CFAR),A*STAR, Singapore \and
Institute of High Performance Computing (IHPC), A*STAR, Singapore \and
ChohoTech Inc., Hangzhou, China
}

\maketitle              % typeset the header of the contribution

\renewcommand{\thefootnote}{\fnsymbol{footnote}}
\footnotetext[1]{These authors contributed equally to this work.}

\begin{abstract}
Optical Intraoral Scanners (IOS) are widely used in digital dentistry to provide detailed 3D information of dental crowns and the gingiva. Accurate 3D tooth segmentation in IOSs is critical for various dental applications, while previous methods are error-prone at complicated boundaries and exhibit unsatisfactory results across patients. In this paper, we propose TSegFormer which captures both local and global dependencies among different teeth and the gingiva in the IOS point clouds with a multi-task 3D transformer architecture. Moreover, we design a geometry-guided loss based on a novel point curvature to refine boundaries in an end-to-end manner, avoiding time-consuming post-processing to reach clinically applicable segmentation. In addition, we create a dataset with 16,000 IOSs, the largest ever IOS dataset to the best of our knowledge. The experimental results demonstrate that our TSegFormer consistently surpasses existing state-of-the-art baselines. The superiority of TSegFormer is corroborated by extensive analysis, visualizations and real-world clinical applicability tests. Our code is available at \url{https://github.com/huiminxiong/TSegFormer}.

\keywords{3D tooth segmentation  \and IOS mesh scans \and Transformer }
\end{abstract}

\section{Introduction}
Deep learning is becoming increasingly popular in modern orthodontic treatments for tooth segmentation in intraoral scans (IOS), cone-beam CT (CBCT) and panoramic X-ray~\cite{conebeam,panoramic}. 
Accurate tooth segmentation in 3D IOS dental models is crucial for orthodontics treatment such as diagnosis, tooth crown-root analysis and treatment simulation~\cite{Yuan,kondo2004tooth}. Specifically, tooth segmentation classifies each triangular face of a 3D IOS tooth model with about 100,000 to 400,000 faces and a spatial resolution of 0.008-0.02mm into teeth and gingiva categories, following the Federation Dentaire Internationale (FDI) standard~\cite{herrmann1967completion}.

There are two main categories for tooth segmentation in IOS: conventional methods that handle 2D image projections~\cite{Yamany,kondo2004tooth,5559877} or directly operate on 3D IOS meshes~\cite{WU2014199,Yuan,Zhao,Sinthanayothin}, and deep learning methods that operate on meshes or point clouds~\cite{XuCNN,Lian,TeethGNN,Sun,zhang2021tsgcnet,lian2020deep,Zanjani,TSegNet21101949,jdr,TMI,midl}.
However, many challenges persist. Complicated morphological topology or dental diseases (e.g. crowded or erupted teeth) can lead to unsatisfactory segmentation performance~\cite{jdr}. Additionally, current methods often fail to recognize mesh faces between adjacent teeth or the tooth and gingiva, requiring time-consuming post-processing to refine the noisy boundary segmentation~\cite{XuCNN,lian2020deep,jdr}. Moreover, the state-of-the-art works such as MeshSegNet~\cite{lian2020deep}, TSGCNet~\cite{zhang2021tsgcnet} and DCNet~\cite{jdr} have only been evaluated with a limited amount of data samples and the clinical applicability need to be evaluated with large-scale dataset or in real-world scenarios.

Inspired by the success of transformers in various tasks \cite{attention,vit,han2021transformer,liu2021swin,pct,zhao2021point}, we propose a novel 3D transformer framework, named TSegFormer, to address the aforementioned challenges. 
In particular, the tooth segmentation task on 3D IOSs is formulated as a semantic segmentation task on point clouds sampled from raw IOS meshes. 
We design the 3D transformer with tailored self-attention layers to capture long-range dependencies among different teeth, learning expressive representations from inherently sophisticated structures across IOSs. In addition, we design a multi-task learning paradigm where another auxiliary segmentation head is introduced to assist in delimiting teeth and gingiva. Furthermore, in view of the confusing boundary segmentation, we devise a novel geometry guided loss based on a newly-defined point curvature to help learn accurate boundaries. The network is trained in an end-to-end manner and requires no complicated post-processing during inference, making it appealing to practical applications.

We collect a large-scale, high-resolution and heterogeneous 3D IOS dataset with 16,000 dental models where each contains over 100,000 triangular faces. To the best of our knowledge, it is the largest IOS dataset to date. Experimental results show that TSegFormer has reached 97.97\% accuracy, 94.34\% mean intersection over union (mIoU) and 96.01\% dice similarity coefficient (DSC) on the large-scale dataset, outperforming previous works by a significant margin. To summarize, our main contributions are:

\begin{itemize}
\item We design a novel framework for 3D tooth segmentation with a tailed 3D transformer and a multi-task learning paradigm, aiming at distinguishing the permanent teeth with divergent anatomical structures and noisy boundaries.
\item We design a geometry guided loss based on a novel point curvature for end-to-end boundary refinement, getting rid of the two-stage and time-consuming post-processing for boundary smoothing. 
%\item We further adopt a multi-task learning scheme, where another teeth-gingiva segmentation head is introduced to improve the segmentation performance. % Does this count?
\item We collect the largest ever 3D IOS dataset for compelling evaluation. Extensive experiments, ablation analysis and clinical applicability test demonstrate the superiority of our method, which is appealing in real-world applications.
\end{itemize}

\begin{figure}
\includegraphics[width=\textwidth]{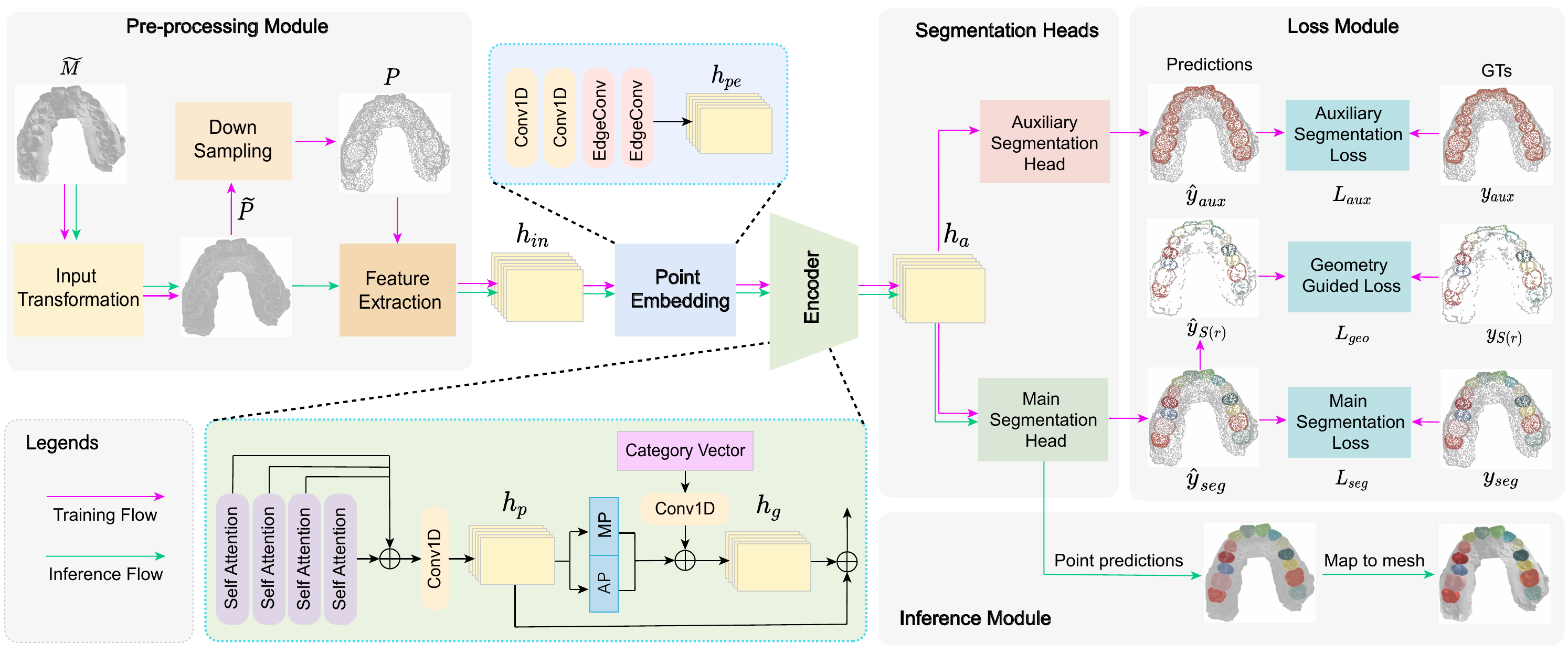}
\caption{The pipeline of our proposed TSegFormer for 3D tooth segmentation
% on digitalized IOS tooth mesh scans.
}
\label{Pipeline}
\end{figure}

\section{Method}
\label{sec 3}
\subsection{Overview}
\label{model overview}
% Given a 3D tooth mesh with $\widetilde{N}$ triangular faces/cells, the target of tooth segmentation is to annotate each mesh cell as one of $C=33$ semantic classes.  Mathematically, for each face $f_i$ in an IOS mesh, we need to assign it a corresponding label $l_i \in \{ 0, 11-18, 21-28, 31-38, 41-48\}$, where $0$ denotes the gingiva and the remaining denote the $32$ permanent tooth classes following the FDI notation. 

The overall pipeline is illustrated in Fig. \ref{Pipeline}. The original mesh $\widetilde{M}$ is converted to a point cloud $\widetilde{P}$ by taking the gravity center point of each mesh face. We downsample a point cloud $P$ with $N=10,000$ points from $\widetilde{P}$, and extract the input feature matrix $h_{in} \in \mathbb{R}^{N \times 8}$ as defined below.
%, which includes the 3D coordinates of points and the normal vector, mean curvature and gaussian curvature of the corresponding mesh faces.  
%Specifically, we first convert each input mesh $\widetilde{M}$ to a point cloud $\widetilde{P}$. During training, we down-sample $\widetilde{P}$ to $N$ points, from which the input feature matrix $h_{in} \in \mathbb{R}^{N \times 8}$ are extracted. 
The network first employs a point embedding module to capture abundant local structure information $h_{pe}$ from $h_{in}$. 
%$h_{in}$ is fed into the point embedding module to help the model learn abundant local structure information $h_{pe}$. 
Thereafter, we design the 3D transformer encoder with self-attention layers to capture high-level semantic representations $h_a$.
% based on $h_{pe}$.
With $h_a$, the main segmentation head produces prediction scores 
% of all $N$ points over 33 classes 
$\hat{y}_{seg} \in \mathbb{R}^{N \times 33}$ (32 permanant teeth and the gingiva), while the auxiliary head generates prediction scores $\hat{y}_{aux} \in \mathbb{R}^{N \times 2}$ to assist distinguishing the tooth-gingiva boundary. 
Furthermore, we devise a geometry guided loss $L_{geo}$, which is integrated with the main segmentation loss $L_{seg}$ and the auxiliary loss $L_{aux}$ to attain superior performance. During inference, we will extract the features $h_{in} \in \mathbb{R}^{\widetilde{N} \times 8}$  for all points in $\widetilde{P}$, process $\widetilde{P}$ into multiple sub-point clouds each with N points, then generate predictions for each point with $\lceil \frac{\widetilde{N}}{N} \rceil$ rounds of inference, and map them back to raw mesh $\widetilde{M}$.

% \begin{figure}
% \includegraphics[width=\textwidth]{img/vis_pe_small.pdf}
% \caption{Segmentation of TSegFormer with different point embedding modules.} 
% \label{pe vis}
% \end{figure}

\subsection{TSegFormer Network Architecture}

\begin{figure}[htbp]
% \centering
\subfigure[Mean cur]{
\label{cur-vis-1}
\includegraphics[width=0.20\linewidth]{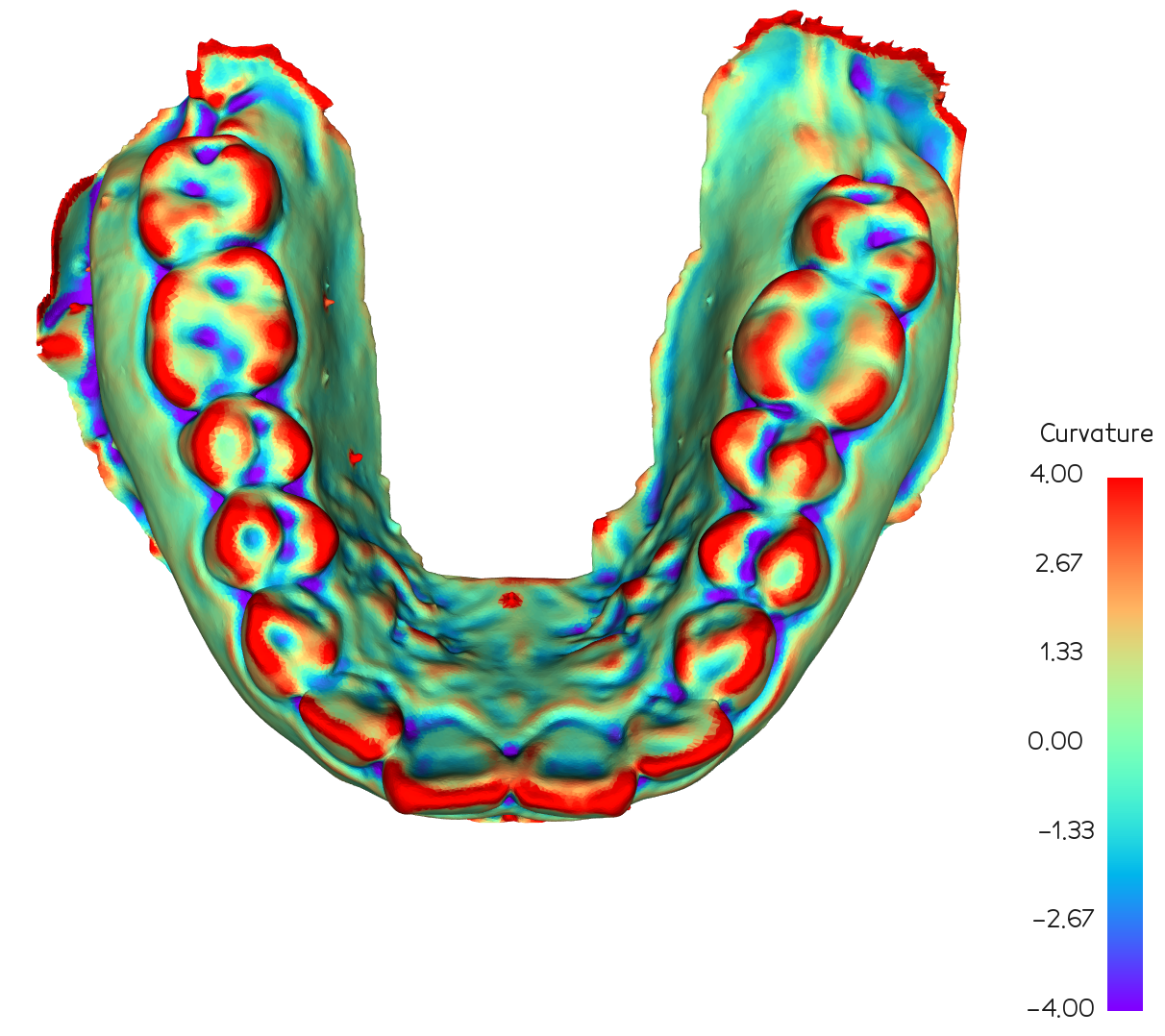}
}
\subfigure[Gaussian cur]{
\includegraphics[width=0.20\linewidth]{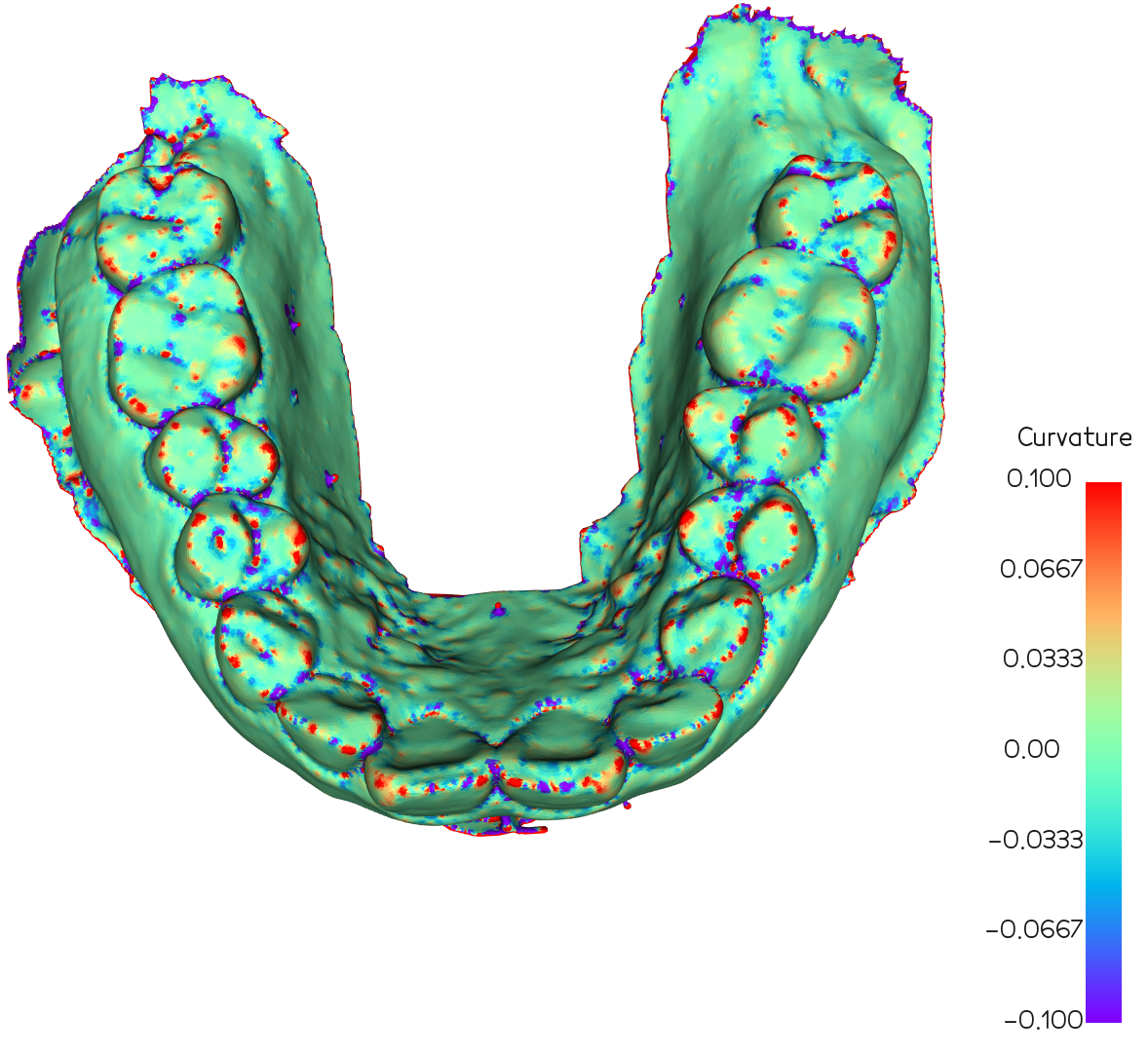}
}
\subfigure[Point cur]{
\label{point cur}
\label{cur-vis-3}
\includegraphics[width=0.20\linewidth]{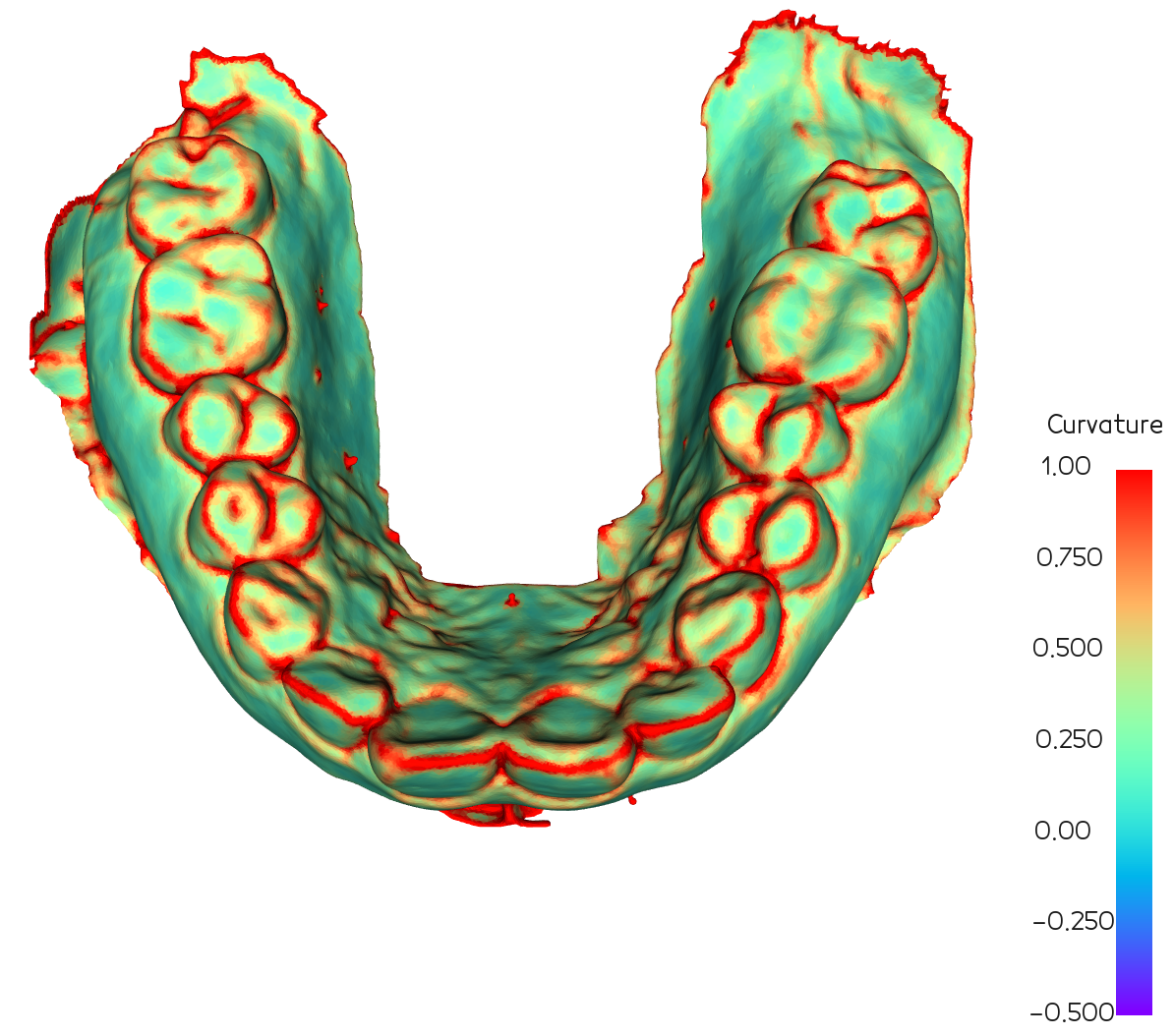}
}
\subfigure[Performance]{
\label{cur res}
\includegraphics[width=0.28\linewidth]{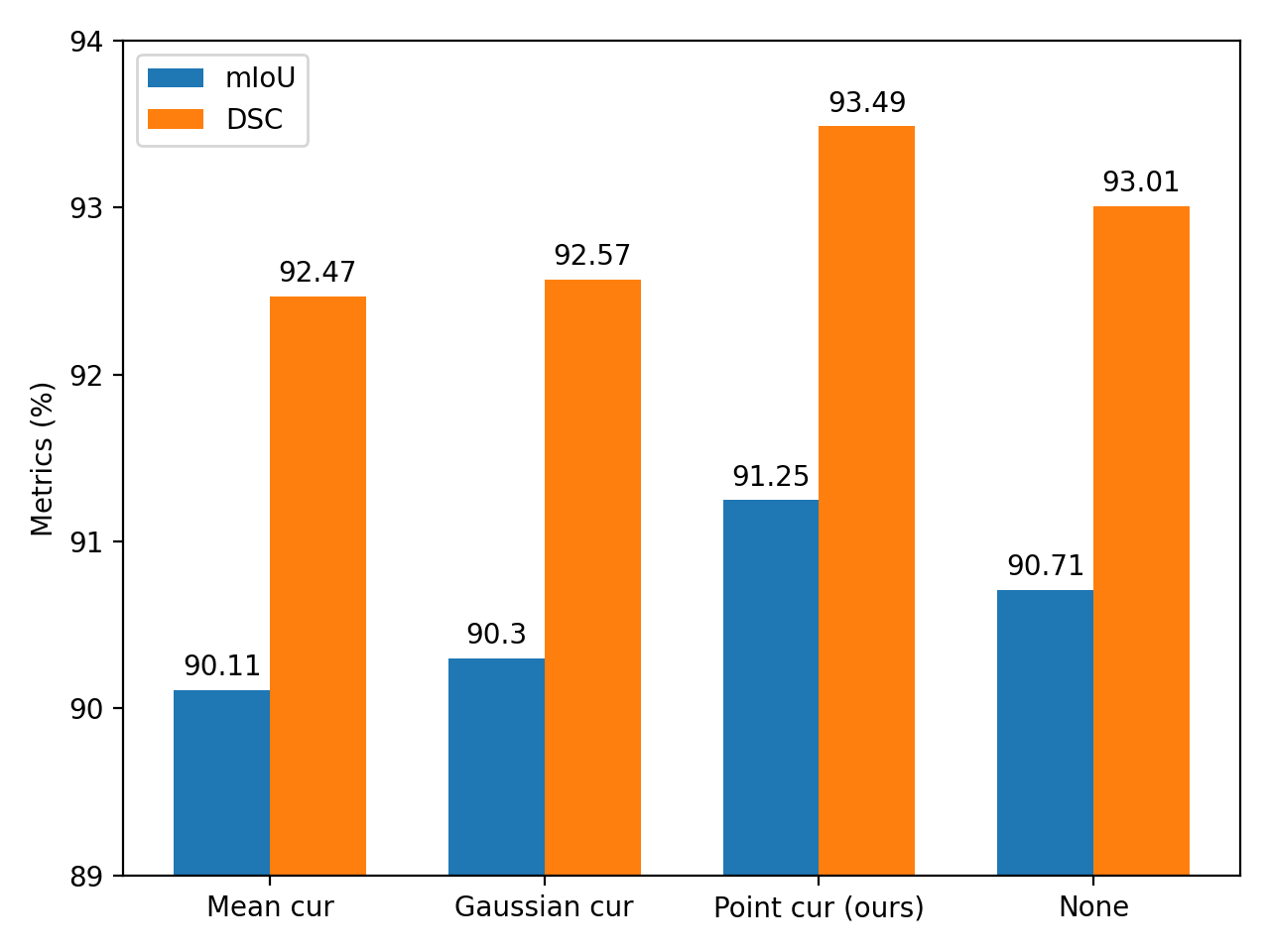}
}
\caption{Visualization and performance comparison of different curvatures(cur).}
\label{vis cur}
\end{figure}

\noindent\textbf{Feature Extraction}
We first transform input meshes to point clouds as directly handling meshes with deep nets is computationally expensive, especially for high-resolution IOSs. To compensate for potential topology loss,
we extract 8-dimensional feature vectors
$h_{in} \in \mathbb{R}^{\widetilde{N} \times 8}$
/$\mathbb{R}^{N \times 8}$ 
for each point to preserve sufficient geometric information, including the point's 3D Cartesian coordinates, 3-dimensional normal vector of mesh face, the Gaussian curvature and a novel point ``curvature'' $m_i$. The $m_i$ is defined as $m_i = \frac{1}{\left| K(i) \right|} \sum_{j \in K(i)} \theta(n_i, n_j)$, 
% \begin{equation}
% \centering
% m_i = \frac{1}{\left| K(i) \right|} \sum_{j \in K(i)} \theta(n_i, n_j), 
% \label{mean curvature}
% \end{equation}  
where $n_i$ is the $i$-th point's 
% or central point's 
normal vector, 
% $n_j$ is the $j$-th point's or a single adjacent point's normal vector, 
$K(i)$ is the second-order neighborhood of the $i$-th point, $\left| K(i) \right|$ is the number of points in $K(i)$, and $\theta(\cdot,\cdot)$ denotes the angle in radians between two vectors. 
By definition, the curvature of a point reflects how much the local geometric structure around this point is curved, i.e., the local geometry on 3D tooth point clouds.

\noindent\textbf{Backbone Network} Delineating complicated tooth-tooth or tooth-gingiva boundaries requires decent knowledge of local geometry in IOS. Hence, we first learn local dependencies from the input $h_{in}$. In particular, we design a point embedding module composed of two linear layers and two EdgeConv layers~\cite{dgcnn}, which takes $h_{in}$ as input and learn local features $h_{pe} \in \mathbb{R}^{N \times d_{e}}$. The point embedding module enriches point representations with local topological information, with ablation results in the Supplementary Material (\textit{SM}) Table 1. %, allowing us to capture decent features in complicated tooth-tooth or tooth-gingiva boundaries. 
Meantime, in view of the inherently sophisticated and inconsistent shapes and structures of the teeth, 
% it's necessary to perceive global relationship by effective representation learning. Given 
and the ability of attention mechanism to capture long-range dependencies and suitability for handling unordered point cloud data \cite{attention,vit,liu2021swin,pct,zhao2021point}, we build an encoder module based on it. 
% to perceive global relationship. 
The encoder module, composed of four successive self-attention layers and a linear transformation, further yields the high-level point feature maps $h_p$.
% shown as
% \begin{equation}
% h_{p} = \sigma \left( h_{sa_1} \oplus h_{sa_2} \oplus h_{sa_3} \oplus h_{sa_4} \right), 
% \end{equation} 
% where  $\sigma(\cdot)$ is a linear layer, and $\oplus$ denotes the channel-wise concatenation.
To avoid misjudging jaw categories, an extra 2D category vector $V$ is fed as input to help distinguish the maxillary and mandible and obtain the global feature maps $h_g$. 
Specifically, 
% \begin{equation}
$h_g = \sigma(V) \oplus MP(h_p) \oplus AP(h_p)$,
% \end{equation}
where ``MP'' and ``AP'' respectively denote the max and average pooling; $\oplus$ denotes concatenation and $\sigma(\cdot)$ is a linear layer.
Finally, we obtain feature maps $h_a$ for all points,  
% by concatenating $h_p$ and $h_g$.
where $h_a = h_p \oplus h_g$. 
% Details settings of the network architecture are in \textit{SM} {\textcolor{red}{Table. 4. }}
% \begin{equation}
% h_a = h_p \oplus h_g.
% \end{equation}

\noindent\textbf{Segmentation Heads}
\label{segmentation heads}
To improve the network's ability to recognize different tooth and gingiva categories, we design two segmentation heads. The main segmentation head, an MLP ($MLP_{seg}$), generates point classification scores for 33 classes $\hat{y}_{seg} =MLP_{seg}(h_a) \in \mathbb{R}^{N \times 33}$ for tooth segmentation. Meanwhile, considering the prevalence of incorrect prediction of tooth-gingiva boundaries, we design an auxiliary segmentation head $MLP_{aux}$ 
% focuses on the boundary between teeth and gum, 
to provide binary classification scores for each point belonging to either tooth or gingiva, i.e., $\hat{y}_{aux} =MLP_{aux}(h_a) \in \mathbb{R}^{N \times 2}$. Experimental results indicate that the cooperation with $MLP_{aux}$ can refine tooth-gingiva segmentation boundary.

\noindent\textbf{Geometry Guided Loss}
Previous methods are usually unsatisfactory to delineate the complicated tooth-tooth boundaries. 
% In TSegFormer, we devise a geometry guided loss which focuses on error-prone regions in a geometry-aware manner with negligible extra computations. 
Observing that points with high point curvatures often lie on the upper sharp ends of tooth crowns and the teeth boundaries (Fig. \ref{point cur}), where mispredictions usually occur, 
% (as illustrated in Fig.  \ref{boundary issues}). 
we define the novel geometry guided loss $L_{geo}$. $L_{geo}$ encourages TSegFormer to adaptively focus more on error-prone points with higher point curvatures with negligible extra computations. Specifically, we define it as
\begin{equation}
\centering
L_{geo} = - \sum _{i \in {S(r)}} \sum_{c=1}^{33}
(1-{\hat{p}}_{ic}^{geo})^{\gamma} \cdot \Phi(y_{S(r)_i}, c) \cdot log(\hat{p}_{ic}^{geo}),
\end{equation}
where $\gamma$ is the modulating factor (empirically set to 2 in experiments); $y_{S(r)_i} \in \mathbb{R}^{33}$ represents the gold label of the $i$-{th} point in the point set $S(r)$; $\hat{p}_{ic}^{geo}$ denotes the predicted probability of the $i$-{th} point belonging to the $c$-{th} class, and $\Phi(y_i, c)$ is an indicator function which outputs 1 if $y_i=c$ and 0 otherwise.
% \begin{equation}
% \Phi(y_i, c)=\left\{
% \begin{aligned}
% 1 & , & if \; y_i=c, \\
% 0 & , & otherwise.
% \end{aligned}
% \right.
% \end{equation}
Concretely, $S(r)$ is a set of points whose point curvatures $m_i$
% (defined in Equation \ref{mean curvature}) 
are among the top $r\cdot 100\%~ (0<r \leq 1)$ of all $N$ points, i.e.,
% \begin{equation}
% $\underbrace{m_{a_1} \geq m_{a_2} \geq \cdots > m_{a_{rN}}}_{S(r):=\{a_1,\ a_2,\ \cdots,\ \lceil a_{rN} \rceil \}} \geq m_{a_{rN+1}} \geq \cdots \geq m_{a_N}.$ 
$S(r):=\{a_1,\ a_2,\ \cdots,\ \lceil a_{rN} \rceil \}$, where $m_{a_1} \geq m_{a_2} \geq \cdots > m_{a_{rN}} \geq m_{a_{rN+1}} \geq \cdots \geq m_{a_N}$.
The experimental results (Fig. \ref{cur res}) on a dataset of 2,000 cases 
indicate that $L_{geo}$ is more effective with 
% universally effective with different curvatures, and moreover, attains better results with 
our point curvature over traditional mean and Gaussian curvatures, even they are worst than no curvature. 
This is because our point curvature provides more clear tooth-tooth and tooth-gingiva boundary indications (Fig. \ref{cur-vis-1}-\ref{cur-vis-3}), thus avoiding misleading the model to focus too much on unimportant non-boundary points.
 
% \end{equation}
% where $a_i,\ a_2,\ \cdots,\ a_N$ are sorted in a decreasing order of point curvature. 
% We denote the prediction scores of these points as $\hat{y}_{S(r)}$, as shown in Fig. \ref{Pipeline}. % Have we ever used this symbol?

%Experimental results indicate $r=0.4$ gives the best performance, as shown in Fig. \ref{point ratio}.

% The geometry guided loss can encourage TSegFormer to catch the hard boundary areas which typically have lower confidence than the upper sharp ends of tooth crowns, and give these hard areas higher specific gravity, similar to the focal loss~\cite{lin2017focal}. 
% Hence, the false predictions in boundary areas can be corrected effectively, enhancing the generalization ability of the network on unseen cases. As shown in the complicated case in Fig. \ref{gingiva loss vis},
% the model with geometry guided loss (Fig. \ref{gingiva loss vis}.(c)) achieves almost perfect recognition for the incisor, while the model without geometry guided loss (Fig. \ref{gingiva loss vis}.(a)) still misclassify part of the incisor as the adjacent lateral incisor.

We employ the  cross entropy loss as the loss of main segmentation head ($L_{seg}$) and the loss of auxiliary segmentation head ($L_{aux}$).
% , which are respectively denoted as $L_{seg}$ and $L_{aux}$.
% $L_{seg} = - \sum _{i=1}^{N} \sum_{c=1}^{33} 
% \Phi(y_{{seg}_i}, c) \cdot log(\hat{p}_{ic}^{seg}),$ and $L_{aux} = - \sum _{i=1}^{N}  y_{{aux}_i} \cdot log(\hat{p}_{i}^{aux}) + (1-y_{{aux}_i}) \cdot log(1 - \hat{p}_{i}^{aux})$,
% % \begin{equation}
% % L_{seg} = - \sum _{i=1}^{N} \sum_{c=1}^{33} 
% % \Phi(y_{{seg}_i}, c) \cdot log(\hat{p}_{ic}^{seg}),
% % % \end{equation}
% % \begin{equation}
% % L_{aux} = - \sum _{i=1}^{N}  y_{{aux}_i} \cdot log(\hat{p}_{i}^{aux}) + (1-y_{{aux}_i}) \cdot log(1 - \hat{p}_{i}^{aux}),
% % \end{equation}
% where $\hat{p}_{ic}^{seg}$ denotes the predicted probability of the $i$-th point belonging to the $c$-th class from the main segmentation head, $\hat{p}_{i}^{aux}$ denotes the predicted probability of the $i$-th point belonging to the gingiva class from the auxiliary head, and $y_{{seg}_i} \in \mathbb{R}^{33}$ and $y_{{aux}_i} \in \mathbb{R}^{2}$ represent the ground truth (GT) labels of the $i$-th point in the main and auxiliary segmentation heads.  
The total loss $L_{total}$ is computed by combining $L_{seg}$, $L_{aux}$ for \textit{all points} and $L_{geo}$ for \textit{hard points}:
% \begin{equation}
 $L_{total} = L_{seg} + \omega_{geo} \cdot L_{geo} + \omega_{aux} \cdot L_{aux}$.
% \end{equation}
% where $\omega_{geo}$ and $\omega_{aux}$ are coefficients. 
We set the weights $\omega_{geo}=0.001$, $\omega_{aux}=1$ and the ratio $r=0.4$, and detailed hyperparameter search results in \textit{SM} Fig. 1 indicate that the performance is stable across different hyperparameter settings.

%Related experiments are conducted as well to demonstrate the effects of different $\omega_{geo}$ and $\omega_{aux}$ on the segmentation performance, in which $\omega_{geo}=0.001$ and $\omega_{aux}=1$ shows the best, as shown in Fig. \ref{geo-guided} and Fig. \ref{aux-branch weight}.

% \begin{figure}[!t]
% \centering
% \subfigure[Geometry-guided loss]{
% \begin{minipage}[t]{0.3\linewidth}
% \centering
% \includesvg[width=\linewidth]{./img/table4.svg}
% % \caption{fig1}
% \label{geo-guided}
% \end{minipage}%
% }%
% \subfigure[Ratios of high-curvature points]{
% \begin{minipage}[t]{0.3\linewidth}
% \centering
% \includesvg[width=\linewidth]{./img/table5.svg}
% %\caption{fig2}
% \label{point ratio}
% \end{minipage}%
% }%
% \subfigure[Auxiliary segmentation loss]{
% \begin{minipage}[t]{0.3\linewidth}
% \centering
% \includesvg[width=\linewidth]{./img/table6.svg}
% %\caption{fig2}
% \label{aux-branch weight}
% \end{minipage}
% }%
% \centering
% \caption{Ablation results for different hyperparamters. (a) Weights of the geometry guided loss. (b) Ratios of the points with high point curvature in the geometry guided loss. (c) Weights of the auxiliary segmentation loss. }
% \end{figure}

\section{Experiments}
\label{sec 4}
\subsection{Dataset and Experimental Setup}  \label{setup}
% \subsubsection{Dataset}
We construct a large-scale 3D IOS dataset consisting of 16,000 IOS meshes with full arches (each with 100,000 to 350,000 triangular faces) collected between 2018-2021 in China, with evenly distributed maxillary and mandible scans labeled by human experts. 
Detailed data statistics are presented in \textit{SM} Table 2, and 39.8\% of the data have third-molars, 16.8\% suffer from missing teeth, which all reveal the complexity of the dataset. 
% Detailed data statistics which reveal the complexity of the dataset, e.g., only 39.8\% of them have third-molars, 16.8\% suffer from missing teeth, are presented in \textit{SM} Table 2. 
The dataset is randomly split into training (12,000 IOSs), validation (2,000 IOSs) and test sets (2,000 IOSs).
% 12,000 for training, 2,000 for validation, and 2,000 for testing. 
%To the best of our knowledge, this is the largest ever 3D IOS dataset.
% for 3D tooth segmentation. 
%\textcolor{blue}{(add contents if clinical analysis is done on val dataset)
Furthermore, we collect an external dataset with 200 complex cases (disease statistics shown in 
\textit{SM} Table 3) to evaluate the real-world clinical applicability of TSegFormer. Detailed training and architecture settings are in \textit{SM} Table 4 and Table 5.

\subsection{Main Results on Tooth Segmentation}
% \paragraph{\textbf{Baselines}}
% To show TSegFormer's superiority over other methods, we trained seven networks under the same dataset and configurations as mentioned in section \ref{setup} as comparison baselines. 

% We conducted comprehensive experiments by comparing our TSegFormer to seven representative and state-of-the-art baselines from three categories: 1) neural networks for point clouds; 2) transformers for point clouds; 3) domain-specific architectures for 3D tooth segmentation. As for traditional neural networks, we chose PointNet++ \cite{pointnet++} and the advanced DGCNN model \cite{dgcnn}, as both of them are widely-used for lots of point cloud tasks. Moreover, we included the point transformer \cite{zhao2021point} and PVT \cite{zhang2021pvt} models as transformer-based baselines, which achieve state-of-the-art performance on various tasks. As for domain specific architectures, we included all the recently proposed deep learning based 3D tooth segmentation architectures whose source codes are publicly available, i.e., MeshSegNet \cite{meshsegnet}, TSGCNet \cite{zhang2021tsgcnet} and DC-Net \cite{jdr}. An interesting finding is that these architectures were evaluated in different settings in their original papers, but not systematically evaluated with large-scale dataset yet. All the baselines were implemented with their original source codes, and followed the same evaluation protocols in \ref{setup} for fair comparison. 

To our best knowledge, there has been no prior work on Transformer-based segmentation on non-Euclidean 3D tooth point clouds/meshes. Hence, % final response 4.1
we compare our TSegFormer to seven representative and state-of-the-art baselines from three categories: 1) neural networks for point clouds, including PointNet++ \cite{pointnet++} and DGCNN \cite{dgcnn}; 2) transformers for point clouds, including point transformer \cite{zhao2021point} and PVT \cite{zhang2021pvt}; 3) domain-specific architectures for 3D tooth segmentation, including MeshSegNet \cite{lian2020deep}, TSGCNet \cite{zhang2021tsgcnet} and DC-Net \cite{jdr}. 
% An interesting finding is that these architectures were evaluated in different settings in their original papers, but not systematically evaluated with large-scale dataset yet. All the baselines were implemented with their original source codes. 
For fair comparison, baselines that cannot achieve raw-resolution mesh prediction followed the same inference protocols in \ref{model overview}, while the rest kept their original inference schemes.

% In terms of point cloud processing, PointNet++ (\citet{pointnet++}) and DGCNN (\citet{dgcnn}) are popular CNN models, and point trasnformer(\citet{zhao2021point}) and PVT(\citet{zhang2021pvt}) are powerful Transformer-based networks designed for general point cloud classification and segmentation tasks. Moreover, MeshSegNet (\citet{meshsegnet}), TSGCNet(\citet{zhang2021tsgcnet}) and DC-Net (\citet{jdr}) are previous state-of-the-art methods for 3D tooth segmentation. We directly replace the TSegFormer network architecture with PointNet++, DGCNN, point transformer and PVT in the training stage to produce the corresponding pretrain models for subsequent inference as described in section \ref{infer}. While we implement DCNet exactly following its original pre-processing, architecture implementation and inference manners, we discard the graph cut based smoothing in canary module for fair comparison. For MeshSegNet and TSGCNet, we also keep the original pre-processing and architecture design but adopt the same reasoning approaches as DCNet to yield the predicted results of all triangular faces in a mesh.

% \paragraph{\textbf{Results}}
% The quantitative segmentation results of TSegFormer and the baselines are presented 
% We can make the following observations. First, TSegFormer obtained an overall accuracy of 97.97\%, mIoU of 94.34\%, DSC of 96.01\%,
We can firstly observe that TSegFormer outperforms existing best-performing point transformer model \cite{zhao2021point} by 0.16\% in accuracy, 1.04\% in mIoU and 0.71\% in DSC (Table \ref{comparison with baseline}). Such an improvement is surely significant considering the complicated real-world cases in our large-scale dataset and the relatively high performance of point transformer with an mIoU of 93.30\%. Moreover, TSegFormer consistently surpassed all baselines on both mandible and maxillary in terms of all metrics, demonstrating its universal effectiveness.

\begin{table}[t]
\centering
\caption{Main segmentation results (Tested on 1,000 patients). }
% "Acc": overall point-level classification accuracy; "mIoU": mean intersection over union; "DSC": dice similarity coefficient (DSC) on all points.}
\label{comparison with baseline}
\centering
\resizebox{1\textwidth}{!}{%
\begin{tabular}{c|ccc|ccc|ccc}
\hline
\multirow{2}{*}{Method} &
  \multicolumn{3}{c|}{Mandible} &
  \multicolumn{3}{c|}{Maxillary} &
  \multicolumn{3}{c}{All} \\ \cline{2-10} 
            &  mIoU $\uparrow$  &  DSC $\uparrow$   &  Acc $\uparrow$  &  mIoU $\uparrow$ &  DSC $\uparrow$  &  Acc $\uparrow$  &  mIoU $\uparrow$ &  DSC $\uparrow$ &  Acc $\uparrow$ \\
            % & (95\% CI)  & (95\% CI)   & (95\% CI)   & (95\% CI)  & (95\% CI)   & (95\% CI)   & (95\% CI)  & (95\% CI)  & (95\% CI)  \\
            \hline\hline
% PointNet    &        &       &       &        &       &       &        &       &      \\
PointNet++  & 81.11 & 85.33 & 94.96 & 83.89 & 87.12 & 96.28 & 82.57 & 86.27 & 95.65\\
            % &  (80.51, 81.72)   &  (84.76, 85.89)   &  (94.69, 95.22)   &  (83.38, 84.39)   &  (86.64, 87.59)   &  (96.07, 96.48)   &  (82.18, 82.97)   &  (85.90, 86.63)   &  (95.48, 95.82) \\ 
DGCNN       & 92.41 & 94.49 & 97.68 & 93.82 & 95.61 & 98.01 & 93.15 & 95.08 & 97.85 \\
            % &  (91.97, 92.85)   &  (94.07, 94.91)   &  (97.50, 97.86)   &  (93.44, 94.20)   &  (95.25, 95.97)   &  (97.87, 98.15)   &  (92.86, 93.44)   &  (94.80, 95.35)   &  (97.74, 97.97) \\ 
            \hline 
point transformer  & 92.61 & 94.83 & 97.55 & 93.93 & 95.72 & 98.06 & 93.3 & 95.3 & 97.81 \\
            % &  (92.14, 93.09)   &  (94.37, 95.28)   &  (97.32, 97.77)   &  (93.56, 94.30)   &  (95.37, 96.07)   &  (97.92, 98.19)   &  (93.01, 93.60)   &  (95.01, 95.58)   &  (97.69, 97.94) \\ 
PVT         & 90.66 & 93.59 & 96.64 & 92.46 & 94.72 & 97.44 & 91.6 & 94.19 & 97.06 \\
            % &  (90.16, 91.15)   &  (93.14, 94.04)   &  (96.44, 96.84)   &  (92.05, 92.87)   &  (94.34, 95.11)   &  (97.29, 97.59)   &  (91.28, 91.92)   &  (93.89, 94.48)  &  (96.93, 97.18) \\ 
            \hline 
% MeshSegNet  & 79.52 & 85.06 & 90.67 & 83.15 & 87.89 & 92.66 & 81.43 & 86.54 & 91.72 \\
MeshSegNet & 82.21 & 86.55 & 91.98 & 85.37  & 89.28 & 93.72 & 83.87 & 87.98 & 92.90 \\
            % &  (78.78, 80.27)   &  (84.35, 85.76)   &  (90.26, 91.08)   &  (82.53, 83.77)   &  (87.31, 88.47)   &  (92.38, 92.94)   &  (80.94, 81.91)   &  (86.09, 87.00)    &  (91.47, 91.96) \\ 
TSGCNet     & 80.71 & 85.23 & 92.78 & 80.97 & 85.28 & 93.86 & 80.85 & 85.25 & 93.34 \\
           % &   (79.90, 81.52)  &   (84.48, 85.98)   &   (92.38, 93.17)   &   (80.18, 81.77)   &   (84.55, 86.01)  &   (93.49, 94.22)  &   (80.28, 81.42)  &   (84.73, 85.78)   &   (93.07, 93.61) \\ 
DCNet       & 91.18 & 93.89 & 97.11 & 92.78 & 95.18 & 97.44 & 92.02 & 94.57 & 97.28 \\
            % &  (90.76, 91.59)   &  (93.49, 94.29)   &  (96.94, 97.27)   &  (92.41, 93.14)   &  (94.83, 95.53)   &  (97.30, 97.57)   &  (91.74, 92.29)   &  (94.31, 94.84)   &  (97.17, 97.39) \\ 
            \hline\hline 
TSegFormer (Our)     & \textbf{93.53}  & \textbf{95.36}  & \textbf{97.72}  & \textbf{95.07}  & \textbf{96.60}  & \textbf{98.20}  & \textbf{94.34}  & \textbf{96.01}   & \textbf{97.97} \\
            % &  (93.04, 94.02)   &  (94.88, 95.84)   &  (97.47, 97.97)   &  (94.70, 95.44)   &  (96.25, 96.96)   &  (98.05, 98.35)   &  (94.03, 94.64)   &  (95.72, 96.31)   &  (97.83, 98.11) \\ 
            \hline
\end{tabular}%
}
\end{table}

\begin{table}[t]
\centering
\caption{Ablation study on different main components.}
\resizebox{\textwidth}{!}{%
\begin{tabular}{cc|ccc|ccc|ccc}
\hline
\multicolumn{2}{c|}{Component} & \multicolumn{3}{c|}{Mandible} & \multicolumn{3}{c|}{Maxillary} & \multicolumn{3}{c}{All} \\ \hline
Geometry guided loss&  Auxiliary branch & mIoU & DSC  & Acc   & mIoU & DSC  & Acc   & mIoU & DSC  & Acc \\ \hline\hline
                    &                   &92.19 &94.27 &97.35  &94.02 &95.69 &97.96  &93.15 &95.01 &97.67  \\
                    &    \CheckmarkBold &92.53 &94.57 &97.43  &94.45 &96.10 &98.02  &93.54 &95.37 &97.74 \\
      \CheckmarkBold&                   &92.46 &94.47 &97.37  &94.45 &96.12 &98.02  &93.51 &95.33 &97.71  \\
      \CheckmarkBold&    \CheckmarkBold &\textbf{92.95} &\textbf{94.88} &\textbf{97.57}  &\textbf{94.46} &\textbf{96.07} &\textbf{98.14}  &\textbf{93.77} &\textbf{95.51} &\textbf{97.87}\\ \hline
\end{tabular}%
}
\label{ablation-geo&aux}
\end{table}
%\XSolidBrush

It is important to integrate advanced architectures with domain-specific design for
% to obtain 
superior performance. We can notice that though MeshSegNet, TSGCNet, and DCNet are all domain-specific 3D tooth segmentation models, their performance, though on par with PVT and DGCNN, is worse than the point transformer. This is also consistent with the superior performance of transformer-based models on standard point cloud processing tasks, which could be mainly attributed to the larger dataset and powerful attention mechanism that better capture global dependencies. Hence, though models like MeshSegNet adopt some task-specific designs to achieve good performance, they still lag behind point transformer when a huge amount of data samples are available. In contrast, our TSegFormer employs the attention mechanism for point representation learning, and meanwhile, adopted task-specific architectures and geometry guided loss to further boost the performance. More statistical results are in \textit{SM} Table 2.

\subsection{Ablation Studies}
%To further investigate the effectiveness of our designs, we conduct additional ablation studies on the crucial components of our model. To be specific, the comparison between using geometry guided loss and the auxiliary branch is summarized in Table \ref{ablation-geo&aux}. In a nutshell, both geometry guided loss as well as the auxiliary branch boost the segmentation performance, and the presence of each of them will not negatively affect the improvement on the performance brought by the other. Additionally, we also tested the effectiveness of using different point embedding techniques. Details are discussed below.

\noindent \textbf{Effectiveness of Geometry Guided Loss}
Table \ref{ablation-geo&aux} shows that introducing the geometry guided loss can improve the performance under all three metrics, e.g., around 0.4\% improvement in mIoU.
% mIoU gets around a 0.4\% improvement.
% These improvements are also consistent with the better visualizations as illustrated in Fig. \ref{vis base gingiva loss both gt}. We can notice that the isolated mispredictions in the figures of the third column are greatly corrected compared to the first column, indicating the significance of perceiving the boundary regions with this loss. 
Besides, we show the universal effectiveness of the geometry guided loss 
% for other architectures 
by adding it to DCNet \cite{jdr}. The performance of DCNet is also enhanced by 1.43\% in mIoU (Table \ref{DCNet+loss}) with this additional loss.
% We conducted an ablation study on the weights of geometry guided loss, as reported in Fig.\ref{geo-guided}.  Notably, all Acc, mIoU, and DSC have similar trends as $\omega_{geo}$ varies. And the highest performance is achieved under all three metrics when $\omega_{geo} = 0.001$, where the model can obtain an Acc of 97.92\%, mIoU of 93.93\%, and DSC of 95.70\%. In addition, we summarize the result of using different ratios of high-curvature in Fig.\ref{point ratio}. The line chart shows that TSegFormer achieves the best performance when $r=0.4$. 
% \textcolor{red}{Overall, the performance is not sensitive to the changes of $\omega_{geo}$ and $r$ in a certain range. }

% \begin{figure}[htbp]
%   \centering
%   \includesvg[width=0.8\linewidth]{./img/table5.svg}
%   \caption{Segmentation performance under different ratios of high-curvature points in geometry-guided loss}
%   \label{point ratio}
% \end{figure}

% \begin{figure}[htbp]
%   \centering
%   \includesvg[width=0.8\linewidth]{./img/table6.svg}
%   \caption{Segmentation performance under different weights of auxiliary branch loss}
%   \label{aux-branch weight}
% \end{figure}

\begin{table}[t]
\centering
\caption{Segmentation performance of DCNet \cite{jdr} with our geometry guided loss.}
\label{DCNet+loss}
\resizebox{0.9\textwidth}{!}{%
\begin{tabular}{c|ccc|ccc|ccc}
\hline
\multirow{2}{*}{Model} & \multicolumn{3}{c|}{Mandible} & \multicolumn{3}{c|}{Maxillary} & \multicolumn{3}{c}{All} \\ \cline{2-10} 
                & mIoU & DSC  & Acc  & mIoU & DSC  & Acc  & mIoU & DSC  & Acc  \\ \hline\hline
DCNet           &87.72 &91.00 &95.99 &90.77 &93.50 &96.87 &89.32 &92.31 &96.46   \\
DCNet+$L_{geo}$ &\textbf{89.66} &\textbf{92.58} &\textbf{96.46} &\textbf{91.75} &\textbf{94.28} &\textbf{97.14} &\textbf{90.75} &\textbf{93.47} &\textbf{96.82} \\ \hline
\end{tabular}%
}
\end{table}

\noindent \textbf{Effectiveness of the Auxiliary Segmentation Head}
%Similar to the geometry guided loss, introducing an auxiliary branch to our model can improve the segmentation performance for the Acc, mIoU and DSC. 
The auxiliary segmentation head is designed to rectify the inaccuracy brought by mislabeling teeth and gingiva near their boundaries. Adding a loss for the auxiliary branch leads to about 0.4\% mIoU performance improvement (Table \ref{ablation-geo&aux}).

\begin{table}[t]
\centering
\caption{Segmentation performance of TSegFormer under different training set scales.}
\label{training set scale}
\resizebox{\textwidth}{!}{
\begin{tabular}{c|ccc|ccc|ccc}
\hline
\multirow{2}{*}{Training set scale} &
  \multicolumn{3}{c|}{Mandible} &
  \multicolumn{3}{c|}{Maxillary} &
  \multicolumn{3}{c}{All} \\ \cline{2-10} 
        & mIoU & DSC  & Acc  & mIoU & DSC  & Acc   & mIoU & DSC  & Acc \\ \hline \hline
500     &86.36 &89.74 &95.45 &89.22 &91.69 &96.57  &87.86 &90.76 &96.04 \\
1,000   &90.60 &93.13 &96.72 &92.84 &94.77 &97.50  &91.78 &93.99 &97.13 \\
 2,000   &92.15 &94.28 &97.31 &94.05 &95.79 &97.92  &93.15 &95.08 &97.63 \\
4,000   &93.10 &95.05 &97.69 &94.68 &96.28 &98.12  &93.93 &95.70 &97.92 \\
8,000   &93.27 &95.15 &97.67 &94.83 &96.40 &98.16  &94.09 &95.81 &97.92 \\
12,000  &\textbf{93.53} &\textbf{95.36} &\textbf{97.72}  &\textbf{95.07} &\textbf{96.60} &\textbf{98.20}  &\textbf{94.34} &\textbf{96.01} &\textbf{97.97}\\ \hline
\end{tabular}
}
\end{table}

\begin{table}[t]
\centering
\caption{Clinical applicability test on the external IOS dataset (200 cases). 
% ``\#" denotes the number of corresponding cases. 
``\#success''/``\#fail'': number of segmentation that meets/does not meet the clinical criteria. \#param: number of parameters in the network. Inf-T: inference time for 200 cases. }
\label{clinical analysis}
% \begin{tabular}{l|l|l|l}
% \hline
% \#cases              & \#cases meeting clinical criteria & \#cases not meeting clinical criteria & clinical error rate(\%)    \\ \hline
% \multicolumn{1}{c|}{200} & \multicolumn{1}{c|}{152}          & \multicolumn{1}{c|}{48}               & \multicolumn{1}{c}{24.0\%} \\ \hline
% \end{tabular}
% \usepackage{graphicx}

\begin{tabular}{c|c|c|c|c|c}
\hline
  Model &
  \multicolumn{1}{c|}{\#success $\uparrow$} &
  \multicolumn{1}{c|}{\#fail $\downarrow$} &
  \multicolumn{1}{c|}{clinical error rate (\%) $\downarrow$} &
  \multicolumn{1}{c|}{\#param $\downarrow$} &
  \multicolumn{1}{c}{Inf-T(s) $\downarrow$} \\ \hline\hline
%\multicolumn{1}{c|}{MeshSegNet}        & 1   & 199 & 99.5 & 1.81M & 182.79 \\
\multicolumn{1}{c|}{MeshSegNet}        & 65  & 135 & 67.5 & 1.81M & 128.56 \\
\multicolumn{1}{c|}{TSGCNet}           & 15  & 185 & 92.5 & 4.13M & 31.40  \\
\multicolumn{1}{c|}{Point Transformer} & 97  & 103 & 51.5 & 6.56M & 437.21 \\
\multicolumn{1}{c|}{DCNet }            & 109 & 91  & 45.5 & \textbf{1.70M} & \textbf{5.79}   \\ \hline
\multicolumn{1}{c|}{TSegFormer (Our)}     & \textbf{152} & \textbf{48}  & \textbf{24.0}   & 4.21M & 23.15 \\ \hline
\end{tabular}%
\end{table}

% \noindent \textbf{Effectiveness of Point Embedding}
% The result of adopting different point embedding strategies is summarized in Table \ref{point-embed}. Apparently, embedding with purely MLP-based structures gives the worst result, 
% %which is logically obvious,
% since they are not able to capture sufficient local contextual information. With the EdgeConv module, the segmentation performance is greatly improved for all three metrics. 
% % Among them, the mIoU and DSC gain about 2\% performance improvement. 
% Besides, we find that combining MLP-based point embedding with EdgeConv can further enhance the performance. 
% %makes the performance even better. 

\noindent \textbf{Effectiveness on Training Data Efficiency} In real-world orthodontic applications, large-scale training data may not be directly accessible due to privacy concerns. Therefore, to show our model's data efficiency, we train our model on datasets with different sizes (Table \ref{training set scale}). With only 500 training samples, TSegFormer is able to 
% reach an overall accuracy of 96.04\%, mIoU of 87.86\%, DSC of 90.76\%, 
surpass PointNet++, MeshSegNet and TSGCNet trained on 12,000 samples. Furthermore, TSegFormer trained with only 2,000 samples can almost outperform all previous models trained on 12,000 samples. Overall, these results demonstrate the exceptional data efficiency of our TSegFormer.

\noindent \textbf{Effectiveness of Local Point Embedding}
\textit{SM} Table 1 shows purely MLP-based structures perform worst due to the lack of local contexts, while EdgeConv layers can make up for this, and the collaboration of both performs best.

% \begin{figure}[!t]
% \centering
% \includegraphics[width=\linewidth]{img/vis_limit_small.pdf}
% \caption{Limitations of our model.}
% \label{vis_limit}
% \end{figure}

\begin{figure}[!t]
\centering
\includegraphics[width=\linewidth]{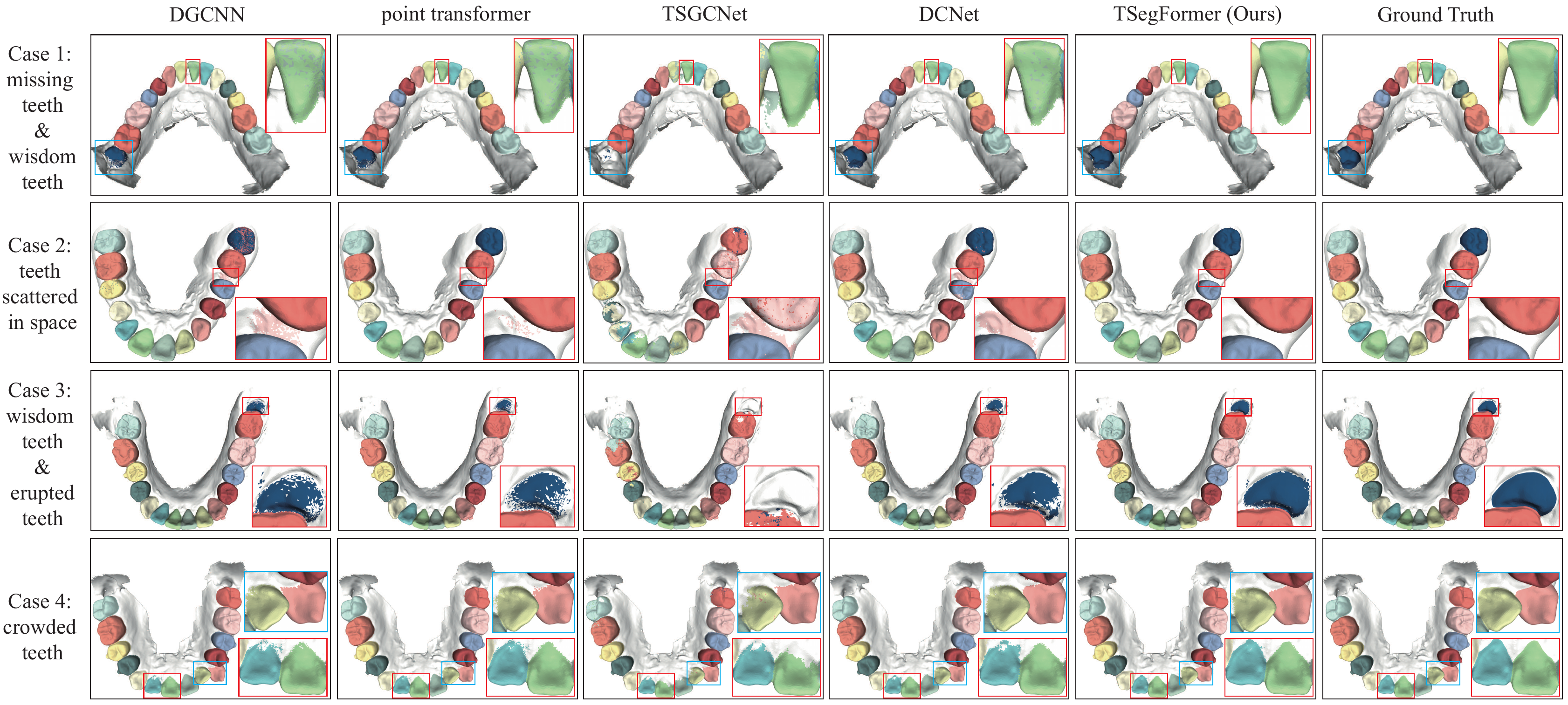}
\caption{Visualization of segmentation of different methods across different diseases.}
\label{vis of different methods}
\end{figure}

\subsection{Clinically Applicability Test and Visualization}
\label{visualization}

%To demonstrate the effectiveness of TSegFormer, we conducted clinical analysis using different models on a validation dataset consisting of 200 cases. The result is shown at Table \ref{clinical analysis}. We visualized the prediction from 5 different models and the visualization is evaluated by a committee of dentists with more than 5-year's experience. As we can see from Table \ref{clinical analysis} that TSegFormer far outperforms the other 4 models regarding the clinical error rate. \textcolor{blue}{The feedback from the dentists indicates that most models cannot meet with the requirement when dealing with the boundary between teeth and gingiva, while TSegFormer apparently handles it better. As for the number of parameters in our network and inference time, though TSegFormer has the second most parameters among all methods we tested, it is the second fastest method that only takes around 23 seconds to complete e2e inference.}

To show the effectiveness of TSegFormer in real-world scenarios, we conducted a clinical applicability test (Table \ref{clinical analysis}) on a dataset with 200 complex IOS scans, whose diseases statistics are in \textit{SM} Table 3. The segmentation with five different models were evaluated by a committee of dentists with more than 5-year experience.  
We can notice that TSegFormer significantly outperforms the other models regarding the clinical error rate.
% i.e., only 24\% of them are not clinically applicable. 

The feedback from dentists indicates that models such as TSGCNet cannot meet the requirement when dealing with complicated boundaries, while TSegFormer apparently handles them better. The point transformer and DCNet also showed promising  performance, but they are yet far behind our TSegFormer. As for the number of parameters and inference time, though TSegFormer has the second most parameters among all methods we tested, it is the second fastest method that only takes around 23 seconds to complete inference for 200 cases, which is certainly acceptable in real-world clinical scenarios.

By visualization, we show the superiority of TSegFormer on various complicated dental diseases in Fig. \ref{vis of different methods}. 
% (e.g. missing teeth, third-molars, erupted teeth, crowded teeth etc), 
% The visualized diseases include the missing teeth, third-molars, dentural diastema, erupted teeth, crowded teeth etc., which usually lead to complicated dentitions and cause difficulties for segmentation in clinical applications. 
The baselines unavoidably produce false predictions or even fail to identify an entire third-molar, while TSegFormer can yield more accurate segmentation and smoother boundaries (see \textit{SM} Fig. 2 for details),
corroborating great potential for clinical applications. Specifically, \textit{SM} Fig. 3 shows that with our geometry guided loss and auxiliary head, the isolated mispredictions and boundary errors are greatly reduced. 
However, TSegFormer fails in some complex samples, e.g. the missing tooth, the erupted wisdom tooth and sunken gingiva and alveolar bone regions, as illustrated in \textit{SM} Fig.4, which needs to be further studied.

\section{Conclusion}
\label{sec 5}
% In this paper, we propose a 3D Transformer-based model, termed TSegFormer, towards more clinically applicable tooth segmentation on large-scale and high-resolution 3D IOS data. With the point embedding module and attention mechanism, TSegFormer can effectively capture both local and global features to distinguish teeth and gingiva in complicated real-world cases. Besides, we propose a novel geometry guided loss based on curvature information to handle the error-prone boundaries, and devise multi-task segmentation heads to further improve the boundary recognition accuracy. The method is evaluated on a large-scale dataset with 16,000 IOS scans, the largest IOS dataset to the best of our knowledge. Comprehensive experiments and clinical applicability tests show that TSegFormer obtains state-of-the-art performance and exhibits superior capability in real-world clinical applications. Our work corroborates the great potential of cutting-edge deep learning techniques in future digital dentistry. \\

% In this paper,
We propose TSegFormer, a 3D transformer-based model for high-resolution IOS segmentation. It combines a point embedding module and attention mechanism to effectively capture local and global features,
% for distinguishing teeth and gingiva in complex clinical cases. 
and introduces a geometry guided loss based on a novel point curvature to handle boundary errors and multi-task segmentation heads for boundary refinement.
% , the largest IOS dataset to the best of our knowledge. 
Results of comprehensive experiments on a large-scale dataset and clinical applicability tests demonstrate TSegFormer's state-of-the-art performance and its great potential in digital dentistry.

\subsubsection*{Acknowledgements.} This work is supported by the National Natural Science Foundation of China (Grant No. 62106222), the Natural Science Foundation of Zhejiang Province, China (Grant No. LZ23F020008) and the Zhejiang University-Angelalign Inc. R$\&$D Center for Intelligent Healthcare.

% The paper introduces TSegFormer, a 3D Transformer-based model for tooth segmentation on high-resolution 3D IOS data, that uses a point embedding module and attention mechanism to capture local and global features. It also includes a geometry guided loss to handle boundary errors and multi-task segmentation heads for boundary recognition. The model is evaluated on a large-scale dataset of 16,000 IOS scans and is shown to have state-of-the-art performance, indicating its potential in digital dentistry. \\

% \subsubsection{Acknowledgements} Please place your acknowledgments at
% the end of the paper, preceded by an unnumbered run-in heading (i.e.
% 3rd-level heading).

\bibliographystyle{splncs04}
\bibliography{ref}
%





% LNCS Homepage, \url{http://www.springer.com/lncs}. Last accessed 4
% \end{thebibliography}
\end{document}